# Task-Dependent Linguistic Measures in Aphasia Discourse: A Comparative Analysis of Narrative, Descriptive, and Procedural Speech

**Emma Al Kubaisi**
Stockholm University, Stockholm, Sweden

ORCID: https://orcid.org/0009-0003-0689-0710

Corresponding author: Emma Al Kubaisi, emma.alkubaisi@gmail.com

## Abstract

**Background:** Transcript-based measures such as word counts, lexical diversity and disfluency rates are widely used to describe language ability in aphasia. Each measure comes from a specific elicitation task, and studies use different tasks, so values from different tasks may not be comparable.

**Aims:** To test whether discourse tasks change these measures within the same speakers, and whether the measures follow a gradient from Procedural to Descriptive to Narrative discourse.

**Methods & Procedures:** Baseline transcripts from 61 adults with chronic post-stroke aphasia were taken from the Fridriksson-2 corpus in AphasiaBank (POLAR trial, ClinicalTrials.gov NCT03416738). Each speaker produced a Procedural (Sandwich), a Descriptive (Broken Window) and a Narrative (Cinderella) sample in one session. A rule-based pipeline extracted 13 measures, 10 of which were common enough to test. Friedman, Page and Wilcoxon tests with false discovery rate correction were used, with a speaker-grouped Random Forest classifier as a supporting check.

**Outcomes & Results:** Total words and number of different words followed the full gradient, but the step from Procedural to Descriptive was small. For most other measures, Narrative stood apart, while Descriptive and Procedural did not differ. Among the 31 speakers with at least 25 words in every task, filler rate, repetition rate and total disfluency density were higher in Narrative than in both other tasks. The classifier reached an accuracy of 0.61 (always guessing the most common task would give 0.41), mainly by recognizing Narrative samples.

**Conclusions:** Discourse tasks are not interchangeable, and expressing disfluencies per word does not remove the task effect. The task should be reported with every measure, and values should be compared only within the same task.

**Keywords:** aphasia; discourse; task effects; disfluency; automated language analysis

## 1. Introduction

Discourse, or connected speech, shows how people with aphasia use language in a setting close to everyday communication. It reveals word-finding difficulties, self-corrections, repetitions and hesitations that single-word tests miss (Dutta et al., 2025). Analyzing discourse by hand takes a great deal of time, and time is the barrier that clinicians and researchers mention most often (Stark & Dalton, 2024). Automated methods reduce this burden. In AphasiaBank, transcribers mark repetitions, revisions, fillers, word fragments and unintelligible speech with fixed codes of the CHAT transcription format, so these behaviors can be counted automatically (MacWhinney, 2000; MacWhinney et al., 2011; Fromm et al., 2024). Measures extracted in this way are now used, for example, to estimate standard aphasia test scores (Riccardi et al., 2024).

Each measure, however, is extracted from a specific task, and studies use different tasks. Automated fluency and macrolinguistic analyses have used only the Cinderella story (Fromm et al., 2024; Lee et al., 2026), a study of word informativeness used a single picture description (Pittman et al., 2025), and a study of aphasia severity combined three tasks (Riccardi et al., 2024). If the task changes the measures, values from different tasks cannot be compared as if they were equal. A difference between two studies may then reflect the tasks rather than the speakers, and a change between two assessments may be mistaken for a change in language ability. In the same speakers with aphasia, picture description and conversation correlate only moderately or weakly (Leaman & Edmonds, 2023), and pooling samples from different genres may hide difficulties in one of them (Bryant et al., 2016).

Tasks are thought to differ in cognitive-linguistic demand. A procedure, such as explaining how to make a sandwich, is a familiar, scripted routine with a fixed order of steps. Procedures give the lowest lexical diversity of all discourse types (Fergadiotis et al., 2011) and simple, noun-heavy phrasing (Stark & Fukuyama, 2021). In a picture sequence, the pictures stay in view, act as external memory and cue the words that are needed. A story retell is the most demanding task: the speaker must recall a long story without pictures, hold its overall structure in mind (Kintsch & van Dijk, 1978) and link the events in time and cause. Story retelling places a greater load on working memory and executive functioning than picture description (Dutta et al., 2025), and narrative performance in aphasia is linked to verbal executive skills (Dutta et al., 2024). These accounts suggest the order Procedural < Descriptive < Narrative. The order is an assumption, not a measurement, and it combines two dimensions that do not have to line up: how scripted the content is, and whether it must be recalled from memory.

Task demands should also affect disfluency. In typical speakers, disfluencies usually signal problems in planning the message, choosing words or building sentences (Didirková, 2024). Fillers such as "uh" and "um" signal that the speaker expects a delay (Clark & Fox Tree, 2002), and disfluency rises with utterance length and with less structured speaking formats (Oviatt, 1995). Tasks that demand more planning and word retrieval should therefore produce more fillers, revisions and breakdowns.

Earlier work shows that tasks matter. In AphasiaBank, discourse type affected most linguistic measures in speakers with and without aphasia (Stark, 2019), and microstructure was most similar for tasks of the same genre (Stark & Fukuyama, 2021). In Mandarin and Croatian, lexical diversity differed between tasks in typical speakers but not significantly in speakers with aphasia (Deng et al., 2024; Jozipović et al., 2026). Most recently, an analysis of almost 10,000 TalkBank samples with nearly 300 automatically extracted features found that task type shaped linguistic output in every clinical group (Stark & Themistocleous, 2026). That study did not include disfluency measures, and not all participants completed all tasks. Evidence on disfluency is more limited. Repetitions and retracing were more frequent in the Cinderella story than in picture-sequence descriptions, but not than in the Sandwich procedure (Culbertson, 2022). Healthy adults were more disfluent per word in an open-ended answer than in a picture-book narrative (Walters et al., 2026), and in very mild aphasia, automated fluency measures were most reliable over repeated testing in narrative discourse (Stark et al., 2026).

A further problem is sample length. Tasks differ greatly in length, and measures such as total words and the number of different words (NDW) grow with the amount of speech. The moving-average type–token ratio (MATTR) was designed to remove this dependence (Covington & McFall, 2010), and disfluencies can be expressed per 100 words. It is still unclear whether task effects on disfluency remain once length is controlled, and, to our knowledge, no study has tested whether features follow the assumed ordered gradient in speakers who all completed every task. The present study addresses these questions with the Fridriksson-2 corpus, in which the same speakers with aphasia completed a Procedural, a Descriptive and a Narrative task.

## 2. Research Question and Hypotheses

This study asks whether the choice of discourse task changes the linguistic measures extracted from aphasia discourse. It is guided by one research question:

**Research question:** Do discourse tasks with different cognitive-linguistic demands produce distinct linguistic profiles, and do these measures follow a systematic complexity gradient across Procedural, Descriptive, and Narrative discourse?

Based on the literature reviewed in the Introduction, two predictions were made:

- **Output and vocabulary:** Features that reflect the amount and variety of speech (total words, number of different words, and mean length of utterance) will follow the full complexity gradient, Procedural < Descriptive < Narrative.
- **Disfluency and breakdown:** Disfluency and breakdown rates per 100 words (fillers, revisions, and unintelligible speech) will be higher in Narrative than in both Descriptive and Procedural discourse, rather than rising steadily across all three tasks.

In addition, the study explored whether length-corrected lexical diversity (MATTR), total disfluency density, and the other disfluency rates differ between tasks. This analysis was exploratory, and no directional prediction was made.

## 3. Method

This study uses a deterministic natural language processing pipeline to extract linguistic features from discourse transcripts, followed by repeated-measures statistical analysis and a machine learning check.

### 3.1 Data Source

The data were obtained from the Fridriksson-2 corpus, which is archived in the AphasiaBank database within the TalkBank system (MacWhinney et al., 2011). AphasiaBank is a shared multimedia resource for the study of language production in aphasia. The Fridriksson-2 corpus contains transcribed discourse from people with post-stroke aphasia, collected as part of the POLAR (Predicting Outcomes of Language Rehabilitation in Aphasia) clinical trial (ClinicalTrials.gov NCT03416738; Kristinsson et al., 2021, 2023). POLAR included adults aged 21–80 who had used English as their primary language for at least 20 years and had chronic aphasia (at least 12 months after a left-hemisphere ischemic or hemorrhagic stroke). People with severely limited speech output or auditory comprehension (a score of 0–1 on either subtest of the Western Aphasia Battery–Revised [WAB-R]) or a bilateral stroke were excluded. Testing took place at the University of South Carolina and the Medical University of South Carolina; participants who entered the trial after the outbreak of COVID-19 were assessed and treated virtually (Kristinsson et al., 2021). Participants received 3 weeks of phonologically focused and 3 weeks of semantically focused naming therapy, in random order, and were assessed before therapy, between and after the two therapy phases, and 1 and 6 months after therapy (Kristinsson et al., 2021, 2023). At each test point, participants completed three tasks from the standard AphasiaBank discourse protocol: the Broken Window picture sequence, the Cinderella story, and the peanut butter and jelly sandwich procedure. All transcripts follow the CHAT transcription standard (MacWhinney, 2000).

Because the trial involved treatment between test points, only pre-treatment baseline transcripts (Timepoint 1) were used. This avoids mixing task effects with treatment-related change: in the same cohort, Cinderella retells changed after treatment, with gains in mean length of utterance (MLU), words per minute and core lexicon (den Ouden et al., 2026). For the within-speaker analysis, only participants who produced all three tasks at baseline were included, because the Friedman and Page tests require complete sets of observations for each speaker. The corpus description lists 68 participants (Fridriksson, 2023), but transcripts are available for 67. Of these, 64 have a baseline transcript: for one speaker the baseline recordings were too glitchy to transcribe, and two speakers start at the second test point. Three baseline transcripts lack a task (one lacks Broken Window because of a glitched video, one has only Broken Window, and one lacks Cinderella). The final sample therefore included 61 speakers and 183 discourse samples.

The 61 speakers are a subset of the POLAR cohort. Table 1 shows the published characteristics of the cohort (Kristinsson et al., 2023); characteristics of the subset are not reported separately.

**Table 1**

*Published Characteristics of the POLAR Cohort (Kristinsson et al., 2023)*

| Characteristic | Mean (SD) or n (%) | Range |
|---|---|---|
| Age (years) | 60.5 (10.9) | 29–80 |
| Female | 42 (41.2%) | |
| Non-Hispanic White | 75 (73.5%) | |
| Right-handed | 90 (88.2%) | |
| Education (years) | 15.4 (2.3) | 12–20 |
| Time since stroke (months) | 46.2 (48.1) | 12–241 |
| NIH Stroke Scale score | 6.2 (3.7) | 0–16 |
| WAB-R Aphasia Quotient | 60.0 (22.7) | 14.5–93.1 |
| ASRS apraxia of speech severity | 1.6 (1.6) | 0–4 |
| History of diabetes | 23 (22.5%) | |
| History of depression | 24 (23.5%) | |
| Aphasia type: Broca's | 45 (44.1%) | |
| Aphasia type: anomic | 29 (28.4%) | |
| Aphasia type: conduction | 16 (15.7%) | |
| Aphasia type: Wernicke's | 7 (6.9%) | |
| Aphasia type: global | 4 (3.9%) | |
| Aphasia type: transcortical motor | 1 (1.0%) | |

*Note.* The values describe the 102 POLAR participants who completed all therapy procedures and at least one outcome assessment, as reported by Kristinsson et al. (2023). The present sample of 61 speakers is drawn from this cohort. NIH = National Institutes of Health (maximum score 42); WAB-R = Western Aphasia Battery–Revised (maximum score 100); ASRS = Apraxia of Speech Rating Scale (maximum severity score 4).

**Ethics.** This study is a secondary analysis of existing, de-identified transcripts from the Fridriksson-2 corpus in AphasiaBank. The original POLAR study was approved by the institutional review boards of the University of South Carolina and the Medical University of South Carolina, and participants provided informed consent for the original study procedures (Kristinsson et al., 2023). The transcripts were accessed through AphasiaBank under the TalkBank data-access and usage rules. Only the text transcripts were used: no audio or video recordings were accessed, the author had no access to any key linking participant codes to identities, no participants were contacted, and no new data were collected.

### 3.2 Discourse Tasks

The three tasks and their assumed demands are summarized in Table 2. Task instructions follow the standard AphasiaBank protocol (MacWhinney et al., 2011). The three tasks were chosen because they come from three different genres (Riccardi et al., 2024; Stark, 2019; Stark & Fukuyama, 2021), are completed by the same speakers in the same session (each baseline transcript is linked to a single video recording), and differ in the amount of support given to the speaker. All 61 speakers completed the tasks in the same order: Broken Window, then Cinderella, then Sandwich. In this corpus, the Sandwich task was given with a photo of a peanut butter and jelly sandwich, which is not the picture that the standard AphasiaBank protocol offers to speakers who have difficulty with the task (Fridriksson, 2023).

**Table 2**

*Discourse Tasks and Their Assumed Cognitive-Linguistic Demands*

| Task | Genre | Task description | Visual support | Main demands | Assumed demand |
|---|---|---|---|---|---|
| Sandwich | Procedural | Explains how to make a peanut butter and jelly sandwich | A photo of a sandwich | Recall of a familiar, scripted routine; small and repeated vocabulary; fixed order of steps | Lowest |
| Broken Window | Descriptive (picture sequence) | Tells what happens in a sequence of pictures | Pictures stay in view | Describing and linking visible events; content is constrained and cued by the pictures | Intermediate |
| Cinderella | Narrative (story retell) | Looks through a wordless picture book, then retells the story after the book is put away | None (book removed) | Recall of a long story from memory; tracking several characters and episodes; temporal and causal links | Highest |

The ordering in the last column is based on the literature reviewed in the Introduction. It is the hypothesis tested by Page's trend test, not an independent measurement of task difficulty.

### 3.3 Data Preprocessing

The transcripts were stored in CHAT (.cha) format, following the AphasiaBank annotation standard. These files contain several tiers, including participant speech, examiner speech, and metadata.

A Python-based preprocessing pipeline was implemented to isolate and clean the participant speech. The pipeline extracts the participant speech tier (*PAR:) and its continuation lines, and removes non-linguistic material, including examiner speech, media headers, and CHAT metadata, so that all analyses are based only on language produced by the participant.

Each transcript was split into task-specific samples using the @G (gem) markers embedded in the CHAT files. These markers were used to separate the Descriptive (Window), Narrative (Cinderella), and Procedural (Sandwich_Picture) tasks, producing one sample per task per participant. The Cinderella_Intro gem, in which the speaker looks through the book, was not analyzed.

The pipeline was implemented with regular expressions (regex) to ensure consistent, deterministic, and reproducible parsing across all files. Words were counted after removing all CHAT codes in square brackets, fillers (&-uh), word fragments (&+fr), other &-codes such as sounds (&=laughs) and interjections from other speakers, the placeholders xxx, yyy and www, omitted-word codes (0-words), and time stamps.

Special-form markers were removed from the word they belong to (for example, son@k was counted as son), and words joined with an underscore (you_know) were counted as one word. Repeated and revised words were kept in the word count. Utterances marked with the postcode [+ exc], which transcribers use for comments that are not part of the task (for example "I don't know" or "okay"), were left out, as in the default analyses of the CLAN program (MacWhinney, 2000). MLU was computed as the number of words divided by the number of utterances that contain at least one word, so that utterances consisting only of fillers, fragments or unintelligible speech do not count as utterances of length zero.

## 3.4 Automated Feature Extraction

### *3.4.1 Disfluency Detection (Regex Engine)*

To turn transcripts into measurable features, the study used a deterministic, rule-based regex engine. The engine counts the CHAT disfluency codes that transcribers insert into the transcripts. It therefore needs no manual coding beyond standard CHAT transcription, but its output depends on the quality and consistency of that transcription.

The engine extracts eight disfluency types: repetitions, revisions, reformulations, fillers, phonemic fragments, pauses, cut-offs, and unintelligible speech. Each marker is identified with a predefined regex pattern that corresponds to a CHAT convention. The categories and their patterns are listed in Table 3. Disfluencies can be classified in several ways, for example into hesitations and self-repairs, or into stuttering-like and other disfluencies (Penttilä et al., 2022), and terminology differs widely between studies (Didirková, 2024). Following the CHAT conventions means that the categories are defined by shared transcription codes rather than by labels chosen for this study.

**Table 3**

*Regex Patterns for Automated Disfluency Extraction*

<table>
<tr><th>Disfluency Type</th><th>Linguistic / Clinical Term</th><th>Regex Pattern</th><th>Description</th></tr>
<tr><td>Repetitions</td><td>Repetition (retracing without correction)</td><td>\[/\]</td><td>Repeated words or phrases</td></tr>
<tr><td>Revisions</td><td>Self-repair (retracing with correction)</td><td>\[//\]</td><td>Speaker corrects or changes an utterance</td></tr>
<tr><td>Reformulations</td><td>Retracing with reformulation</td><td>\[///\]</td><td>Speaker restarts with a different formulation</td></tr>
<tr><td>Fillers</td><td>Filled pauses</td><td>&-\w+</td><td>Non-lexical fillers (e.g., "uh", "um")</td></tr>
<tr><td>Phonemic fragments</td><td>Word fragments</td><td>&\+\w+</td><td>Incomplete or partial word productions (e.g., &+fr)</td></tr>
<tr><td>Pauses</td><td>Pauses</td><td>\((?:\.{1,3}|[\d:]*\d\.?\d*)\)</td><td>Pause codes (.), (..), (...) and timed pauses such as (1.5)</td></tr>
<tr><td>Cut-offs</td><td>Cut-off words</td><td>\b[A-Za-z]+-(?=\s|$)</td><td>Words cut off abruptly (e.g., "go-")</td></tr>
<tr><td>Unintelligible</td><td>Unintelligible speech</td><td>\b(?:xxx|yyy)\b</td><td>Speech that cannot be understood</td></tr>
</table>

### *3.4.2 Linguistic Metrics*

In addition to the disfluency counts, the pipeline computes measures of speech output, utterance length, lexical diversity, and overall disfluency. The full feature set is summarized in Table 4.

- **Total Word Count (TWC)** measures overall speech output.
- **Mean Length of Utterance (MLU)** is the average number of words per utterance and reflects utterance length and complexity.
- **Number of Different Words (NDW)** counts the unique words in a sample and reflects lexical variety. NDW depends strongly on sample length.
- **Moving-Average Type–Token Ratio (MATTR)** computes the type–token ratio within a moving window of fixed size and averages it across the sample (Covington & McFall, 2010). Unlike NDW, it is largely independent of sample length. A window of 25 words was used, because many Procedural and Descriptive samples were short.
- **Disfluency rates** express each disfluency count per 100 words, so that they do not grow with the amount of speech.
- **Total Disfluency Density (TDD)** is the total number of detected disfluency markers divided by the total word count:

$$TDD = (Disfluency\ Markers) / (Total\ Word\ Count)$$

Rates, total disfluency density and MATTR are not reliable for very short samples. A sample with fewer than 25 words was therefore given no value for these measures, and for each of these measures a speaker was included only if all three samples had at least 25 words. Total word count, NDW, and MLU were computed for all samples.

**Table 4**

*Linguistic Features Analyzed*

| Feature | Domain | Definition | Grows with sample length? |
|---|---|---|---|
| Total Word Count (TWC) | Output | Number of words produced | Yes |
| Mean Length of Utterance (MLU) | Output / syntax | Words per utterance | No (average per utterance) |
| Number of Different Words (NDW) | Lexical diversity | Number of unique words | Yes |
| MATTR | Lexical diversity | Mean type–token ratio in a 25-word moving window | No (by design) |
| Fillers /100 words | Disfluency | Filled pauses per 100 words | No (rate) |
| Repetitions /100 words | Disfluency | [/] markers per 100 words | No (rate) |
| Revisions /100 words | Disfluency | [//] markers per 100 words | No (rate) |
| Unintelligible /100 words | Breakdown | xxx / yyy segments per 100 words | No (rate) |
| Phonemic fragments /100 words | Disfluency | &+ fragments per 100 words | No (rate) |
| Pauses /100 words | Disfluency | Pause codes per 100 words | No (rate) |
| Reformulations /100 words | Disfluency | [///] markers per 100 words | No (rate) |
| Cut-offs /100 words | Disfluency | Cut-off words per 100 words | No (rate) |
| Total Disfluency Density (TDD) | Disfluency | All disfluency markers / TWC | No (rate per word) |

*Note.* Pauses, reformulations and cut-offs were too rare to test (more than 90% of samples had none; see Section 4.1).

### 3.5 Statistical Analysis

Because each speaker produced all three tasks, the design is fully within-subject. This is important because disfluency behavior is highly individual. Even among typical speakers, individuals can be told apart from their disfluency patterns alone (Braun et al., 2023). Comparing each speaker with themselves removes this source of variation. The feature distributions were skewed, and many disfluency counts were low, so non-parametric repeated-measures tests were used. A measure was tested only if fewer than 90% of its samples were zero. The analysis followed five steps.

**Step 1: Friedman test.** For each feature, a Friedman test (Friedman, 1937) tested whether the three tasks differed overall. The Friedman test ranks the three values within each speaker, so it compares the tasks while controlling for differences between speakers. Kendall's W (Kendall & Babington Smith, 1939) was used as the effect size, where 0 means no consistent task effect and 1 means that all speakers rank the tasks in the same way.

**Step 2: Page's trend test.** The Friedman test only asks whether the tasks differ. To test the ordered hypothesis directly, Page's trend test (Page, 1963) was applied with the predicted order Procedural < Descriptive < Narrative. Page's test is more powerful than the Friedman test when the alternative hypothesis specifies an order. A significant Page's test shows a trend in the predicted direction, but it does not show that every step of the order is present.

**Step 3: Pairwise Wilcoxon signed-rank tests.** To locate the differences, each pair of tasks (Narrative vs. Procedural, Narrative vs. Descriptive, and Descriptive vs. Procedural) was compared with a two-sided Wilcoxon signed-rank test (Wilcoxon, 1945). The effect size was the matched-pairs rank-biserial correlation r, which ranges from −1 to 1; a positive value means that the first task of the pair had higher values. Its 95% confidence interval was estimated with a bootstrap over speakers (2,000 resamples).

**Step 4: False discovery rate correction.** Because many features and tests were evaluated, p-values were corrected with the Benjamini–Hochberg false discovery rate (FDR) procedure (Benjamini & Hochberg, 1995), separately for each type of test: all Friedman tests, all Page tests, and all pairwise Wilcoxon tests. Results were considered significant at $q < .05$.

**Step 5: Classification of response patterns.** Based on Steps 1–4, each feature was assigned to one of four patterns, using the rules below in this order:

- **Full gradient:** significant Page's trend and all three pairwise differences significant in the predicted direction (Procedural < Descriptive < Narrative).
- **Narrative-higher:** Narrative significantly higher than both Descriptive and Procedural, with no significant difference between Descriptive and Procedural.
- **No evidence of a task effect:** no significant Friedman effect after FDR correction. Because a non-significant test does not show that two tasks are the same, these features are not described as stable.
- **Other:** any other pattern.

All analyses were carried out in Python 3.14 using SciPy 1.18 (Virtanen et al., 2020) for the statistical tests and scikit-learn 1.9 (Pedregosa et al., 2011) for the classifier.

### 3.6 Machine Learning Validation

The statistical tests examine one feature at a time. As a complementary check, a supervised classifier was used to test whether the features, taken together, carry enough information to identify which task a sample came from. The classifier is not the main experiment. Its role is to provide supporting evidence that the three tasks have distinct linguistic signatures.

A Random Forest classifier (Breiman, 2001) with 500 trees (minimum of 5 samples to split a node, square root of the number of features at each split) was trained to assign each sample to one of the three tasks. The features were total word count, NDW, MLU, total disfluency density and the five tested disfluency rates, used as raw values without scaling. MATTR was left out because short samples have no value, and samples with a missing value were excluded, which left 126 samples from 52 speakers. A second model was trained without total word count and NDW, to see whether the tasks can still be told apart when the amount of speech is not available. No settings were tuned.

Performance was evaluated with 10-fold cross-validation grouped by speaker, so that no speaker appeared in both the training and the test data. Record-wise splits allow a model to recognize individual speakers and can inflate accuracy (Saeb et al., 2017). Performance was reported as overall accuracy, macro-averaged F1-score, and per-class precision, recall, and F1-score, together with the confusion matrix. With three classes, chance-level macro-F1 is about 0.33. Because samples with missing values were excluded, the classes were not balanced (34 Procedural, 40 Descriptive and 52 Narrative samples), so always guessing Narrative would give an accuracy of 0.41.

## 4. Results

### 4.1 Sample and Descriptive Statistics

The final sample included 61 speakers, each contributing one Procedural, one Descriptive, and one Narrative sample (183 samples). After utterances marked [+ exc] were left out (364 of 3,338 utterances), descriptive statistics for each feature and task were computed (Table 5). Median values are reported because most features were not normally distributed.

Many samples were short: 27 Procedural, 21 Descriptive and 9 Narrative samples had fewer than 25 words. Thirty-one speakers had at least 25 words in all three tasks, so the analyses of disfluency rates, total disfluency density and MATTR are based on these 31 speakers. Speakers who produce very short samples, often those with more severe aphasia, are therefore under-represented in these analyses. MLU could be computed for 58 speakers, because three speakers had a task without any utterance containing a word. Among the 126 samples with at least 25 words, pauses were coded in only 11 samples (91% zero), reformulations in one sample, and cut-offs in none. These three measures were therefore not tested.

**Table 5**

*Descriptive Statistics by Task, Median (IQR)*

| Feature | n | Procedural (Sandwich) | Descriptive (Broken Window) | Narrative (Cinderella) |
|---|---|---|---|---|
| Total Word Count (TWC) | 61 | 28 (13–51) | 35 (21–67) | 191 (52–311) |
| NDW | 61 | 19 (10–26) | 22 (11–36) | 81 (27–128) |
| MLU | 58 | 5.80 (4.00–8.00) | 6.56 (3.75–8.11) | 7.27 (4.58–9.02) |
| MATTR | 31 | 0.70 (0.62–0.74) | 0.69 (0.61–0.75) | 0.74 (0.72–0.76) |
| TDD | 31 | 0.13 (0.05–0.24) | 0.11 (0.07–0.19) | 0.20 (0.12–0.28) |
| Fillers /100 words | 31 | 3.61 (1.89–6.59) | 2.86 (0.00–4.89) | 5.56 (3.61–10.43) |
| Revisions /100 words | 31 | 2.50 (0.00–4.06) | 2.99 (1.55–5.56) | 4.73 (2.64–6.94) |
| Repetitions /100 words | 31 | 0.00 (0.00–5.88) | 1.85 (0.00–6.38) | 3.63 (1.82–6.06) |
| Unintelligible /100 words | 31 | 0.00 (0.00–0.00) | 0.00 (0.00–0.00) | 0.00 (0.00–0.82) |
| Phonemic fragments /100 words | 31 | 1.82 (0.00–5.26) | 2.17 (0.00–5.40) | 3.78 (2.04–5.27) |

*Note.* n = number of speakers in each analysis. Disfluency rates, TDD and MATTR include only speakers with at least 25 words in all three tasks. IQR = interquartile range; NDW = number of different words; MLU = mean length of utterance; MATTR = moving-average type–token ratio; TDD = total disfluency density.

**Figure 1**

*Per-Speaker Values Across the Three Tasks*

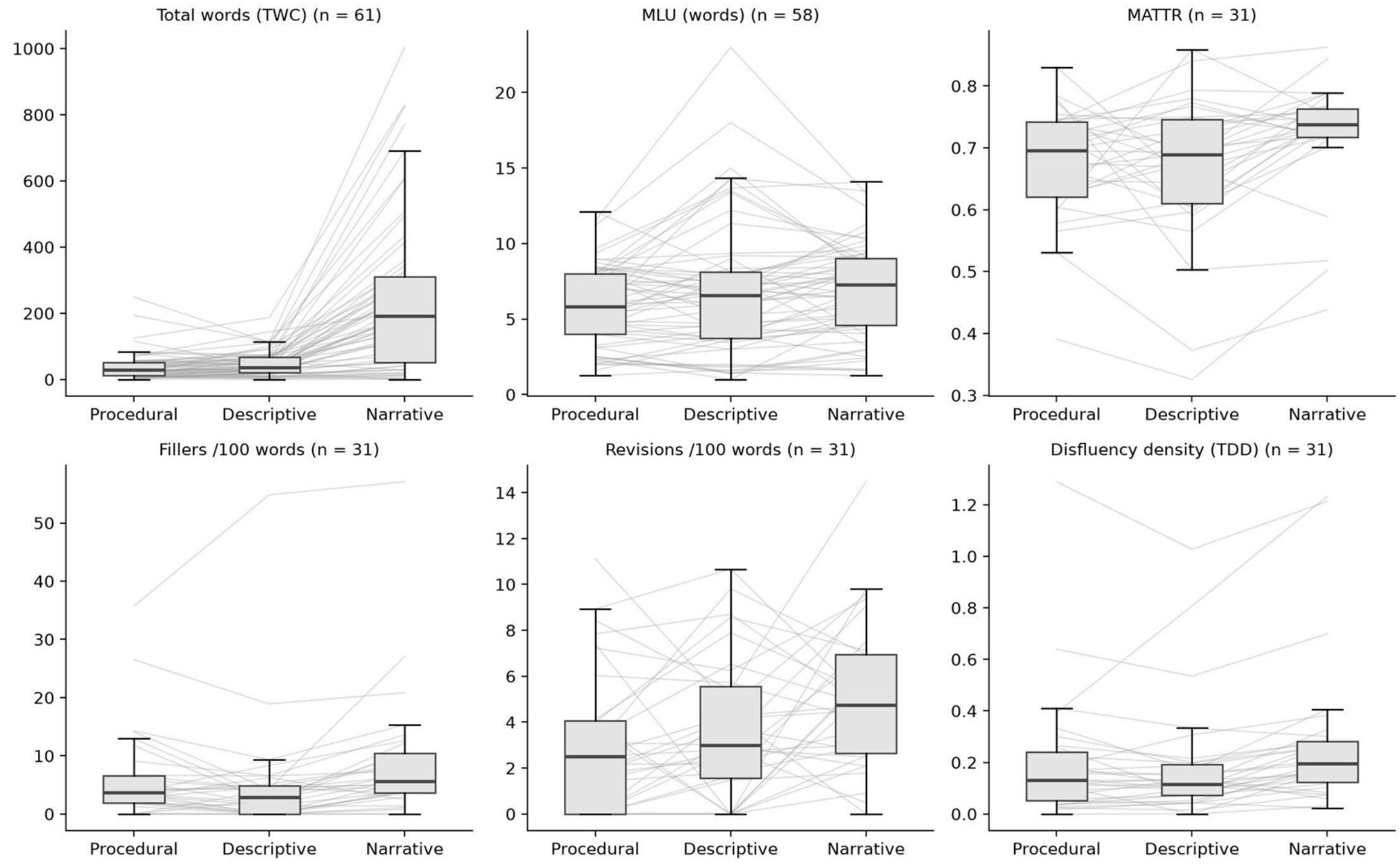

*Note.* Each line connects the three samples from one speaker, and the boxes show the median and interquartile range. Panels show total word count (TWC), mean length of utterance (MLU), moving-average type–token ratio (MATTR), filler rate, revision rate and total disfluency density (TDD). MATTR, the rates and TDD are shown for the 31 speakers with at least 25 words in every task.

## 4.2 Global Task Differences

Table 6 shows the results of the Friedman and Page tests. These are the primary results of the study, because they test directly whether task changes each measure and whether the change follows the predicted order.

**Table 6**

*Friedman and Page's Trend Test Results (Predicted Order: Procedural < Descriptive < Narrative)*

| Feature | n | Friedman $\chi^2(2)$ | q (FDR) | Kendall's W | Page's L | q (FDR) |
|---|---|---|---|---|---|---|
| Total Word Count (TWC) | 61 | 73.23 | < .001* | 0.60 | 821 | < .001* |
| NDW | 61 | 68.00 | < .001* | 0.56 | 817 | < .001* |
| MLU | 58 | 9.93 | .010* | 0.09 | 728 | .003* |
| MATTR | 31 | 9.55 | .011* | 0.15 | 392 | .007* |
| TDD | 31 | 19.93 | < .001* | 0.32 | 400 | < .001* |
| Fillers /100 words | 31 | 22.50 | < .001* | 0.36 | 392 | .007* |
| Revisions /100 words | 31 | 8.69 | .014* | 0.14 | 394 | .004* |
| Repetitions /100 words | 31 | 11.86 | .004* | 0.19 | 395 | .003* |
| Unintelligible /100 words | 31 | 12.12 | .004* | 0.20 | 386 | .048* |
| Phonemic fragments /100 words | 31 | 3.50 | .174 | 0.06 | 383 | .081 |

*Note.* q = p-value after Benjamini–Hochberg false discovery rate (FDR) correction, applied separately to the Friedman tests and to the Page tests. * q < .05. NDW = number of different words; MLU = mean length of utterance; MATTR = moving-average type–token ratio; TDD = total disfluency density.

The Friedman tests showed that task had a significant effect on 9 of the 10 features after FDR correction. The largest effects were found for total word count and NDW (W = 0.60 and 0.56), which means that most speakers ranked the three tasks in the same order on these measures. The effects on the disfluency measures with a significant task effect were moderate (W = 0.14–0.36), and the effects on MATTR and MLU were small (W = 0.15 and 0.09). Page's trend test was also significant for 9 of the 10 features, but as the pairwise tests show, for most of them the trend came from the Narrative task alone. The only feature without a task effect was the phonemic fragment rate (Friedman q = .174).

## 4.3 Pairwise Comparisons

The pairwise Wilcoxon tests (Table 7) show where the task differences lie.

**Table 7**

*Pairwise Wilcoxon Signed-Rank Tests: Effect Size r [95% CI] and FDR-Corrected q*

| Feature | Narrative vs. Procedural | Narrative vs. Descriptive | Descriptive vs. Procedural |
|---|---|---|---|
| Total Word Count (TWC) | r = 0.99 [0.96, 1.00]<br>q < .001* | r = 0.96 [0.89, 0.99]<br>q < .001* | r = 0.37 [0.08, 0.63]<br>q = .029* |
| NDW | r = 0.98 [0.94, 1.00]<br>q < .001* | r = 0.95 [0.88, 0.99]<br>q < .001* | r = 0.36 [0.07, 0.61]<br>q = .029* |
| MLU | r = 0.58 [0.33, 0.79]<br>q < .001* | r = 0.22 [−0.08, 0.53]<br>q = .193 | r = 0.24 [−0.06, 0.51]<br>q = .165 |
| MATTR | r = 0.54 [0.18, 0.82]<br>q = .020* | r = 0.65 [0.35, 0.89]<br>q = .004* | r = −0.04 [−0.43, 0.38]<br>q = .870 |
| TDD | r = 0.69 [0.37, 0.93]<br>q = .002* | r = 0.84 [0.60, 0.98]<br>q < .001* | r = −0.22 [−0.61, 0.22]<br>q = .353 |
| Fillers /100 words | r = 0.53 [0.17, 0.82]<br>q = .020* | r = 0.89 [0.68, 1.00]<br>q < .001* | r = −0.38 [−0.75, 0.01]<br>q = .126 |
| Revisions /100 words | r = 0.59 [0.24, 0.88]<br>q = .013* | r = 0.32 [−0.08, 0.67]<br>q = .172 | r = 0.33 [−0.09, 0.76]<br>q = .182 |
| Repetitions /100 words | r = 0.50 [0.12, 0.80]<br>q = .029* | r = 0.64 [0.31, 0.91]<br>q = .007* | r = 0.23 [−0.25, 0.62]<br>q = .382 |
| Unintelligible /100 words | r = 0.43 [−0.15, 1.00]<br>q = .182 | r = 0.72 [0.22, 1.00]<br>q = .029* | r = −0.33 [−1.00, 0.79]<br>q = .479 |
| Phonemic fragments /100 words | r = 0.18 [−0.22, 0.58]<br>q = .409 | r = 0.47 [0.07, 0.80]<br>q = .041* | r = −0.23 [−0.66, 0.28]<br>q = .394 |

*Note.* r = matched-pairs rank-biserial correlation; a positive value means that the first task of the pair had higher values. 95% confidence intervals from 2,000 bootstrap resamples of speakers. q = p-value after Benjamini–Hochberg correction across all 30 pairwise tests. * q < .05. CI = confidence interval; NDW = number of different words; MLU = mean length of utterance; MATTR = moving-average type–token ratio; TDD = total disfluency density.

The clearest contrast was between Narrative and Procedural discourse. Compared with the Sandwich procedure, the Cinderella retell produced more words (r = 0.99), a higher NDW (r = 0.98), longer utterances (r = 0.58), and, per 100 words, more fillers (r = 0.53), more revisions (r = 0.59) and more repetitions (r = 0.50), as well as a higher total disfluency density (r = 0.69). The unintelligible-speech rate did not differ reliably between these two tasks (r = 0.43, q = .182).

The comparison between Narrative and Descriptive discourse followed the same direction for most features. Cinderella samples were longer than Broken Window samples (r = 0.96) and had higher rates of fillers (r = 0.89), repetitions (r = 0.64), unintelligible speech (r = 0.72) and phonemic fragments (r = 0.47), and a higher total disfluency density (r = 0.84). MLU and the revision rate did not differ reliably between these two tasks (q = .193 and .172).

The comparison between Descriptive and Procedural discourse showed the fewest differences. Broken Window samples contained somewhat more words and a higher

NDW than Sandwich samples (r = 0.37 and 0.36), but no measure that is independent of sample length differed between the two tasks. The filler rate was, if anything, lower in the Descriptive task (r = −0.38, q = .126).

MATTR was slightly higher in Narrative than in Procedural (r = 0.54, q = .020) and Descriptive discourse (r = 0.65, q = .004), with no difference between Descriptive and Procedural (r = −0.04). The narrative retells, although much longer, were therefore not less lexically diverse once sample length was controlled; if anything, they were slightly more diverse.

### 4.4 Which Measures Follow the Complexity Gradient?

Using the decision rules in Section 3.5, each feature was assigned to a response pattern. The results are summarized in Table 8.

**Table 8**

*Response Patterns of Linguistic Measures Across Tasks*

| Pattern | Definition | Features | Interpretation |
|---|---|---|---|
| Full gradient | Procedural < Descriptive < Narrative, all pairs significant | Total Word Count, NDW | The amount of speech rises across the tasks, but most of the rise comes from the Narrative task |
| Narrative-higher | Narrative > Descriptive ≈ Procedural | Fillers /100 words, repetitions /100 words, TDD, MATTR | Per-word disfluency rises, and lexical diversity does not fall, when the speaker must retell a story from memory |
| No evidence of a task effect | Friedman not significant | Phonemic fragments /100 words | Fragment rate did not change reliably with task |
| Other | Any other pattern | MLU and revisions /100 words (Narrative > Procedural only); unintelligible /100 words (Narrative > Descriptive only) | Higher in Narrative, but not clearly different from both other tasks |
| Not tested | More than 90% zeros | Pauses, reformulations, cut-offs | Too rarely coded in these transcripts |

*Note.* NDW = number of different words; MLU = mean length of utterance; MATTR = moving-average type–token ratio; TDD = total disfluency density.

**Full gradient.** Total word count and NDW increased from Procedural to Descriptive to Narrative discourse, with all three steps significant. These measures support the prediction for output and vocabulary. The gradient was, however, very uneven: the step from Sandwich to Broken Window was small (r = 0.37 and 0.36), while the step to Cinderella was very large (r ≥ 0.95). MLU did not follow the full gradient. It was higher in Narrative than in Procedural discourse, but the Descriptive task did not differ reliably from either. The prediction for output and vocabulary was therefore supported for total word count and NDW but not for MLU.

**Narrative-higher.** The filler rate, the repetition rate and total disfluency density were higher in Narrative discourse than in both other tasks, but did not differ between Descriptive and Procedural discourse. Because these measures are expressed per word, the Narrative increase is not simply a consequence of the longer samples. The filler rate supports the prediction for disfluency and breakdown. The other two measures in that prediction gave a partial picture: the revision rate was higher in Narrative than in Procedural discourse only, and the unintelligible-speech rate was higher in Narrative than in Descriptive discourse only. The disfluency prediction was therefore fully supported for one of its three measures and partly supported for the other two. MATTR also met the Narrative-higher rule, with small effects.

**No evidence of a task effect.** The phonemic fragment rate did not change reliably with task overall, although it was higher in Narrative than in Descriptive discourse in the pairwise test.

### 4.5 Machine Learning Validation

The Random Forest classifier was trained to identify the task from the linguistic features. Its performance is shown in Table 9.

**Table 9**

*Random Forest Classification of Discourse Task (10-Fold Cross-Validation Grouped by Speaker)*

| Task | Precision | Recall | F1-Score | Support |
|---|---|---|---|---|
| Procedural (Sandwich) | 0.52 | 0.44 | 0.48 | 34 |
| Descriptive (Broken Window) | 0.49 | 0.53 | 0.51 | 40 |
| Narrative (Cinderella) | 0.76 | 0.79 | 0.77 | 52 |
| Overall accuracy | | | 0.61 | 126 |
| Macro-average | 0.59 | 0.58 | 0.59 | 126 |

*Note.* Features: total word count, number of different words, mean length of utterance, total disfluency density and five disfluency rates per 100 words. Samples with fewer than 25 words are excluded (126 samples from 52 speakers). Chance-level macro-F1 is about 0.33; always guessing Narrative would give an accuracy of 0.41.

The classifier identified the task with an accuracy of 0.61 and a macro-F1 of 0.59. Narrative samples were recognized well (41 of 52), while Procedural and Descriptive samples were often confused with each other: 13 of 34 Procedural samples were labeled Descriptive, and 12 of 40 Descriptive samples were labeled Procedural (Figure 2). Confusions between Procedural and Narrative samples were less common: 6 of 34 Procedural samples were labeled Narrative, and 2 of 52 Narrative samples were labeled Procedural. Without total word count and NDW, accuracy fell to 0.52 and macro-F1 to 0.48. Narrative samples were still recognized (recall 0.79), but Procedural and Descriptive samples were classified close to chance (recall 0.35 and 0.33). No formal test against chance was run, so these results are descriptive.

**Figure 2**

*Confusion Matrix for the Random Forest Classifier (All Features)*

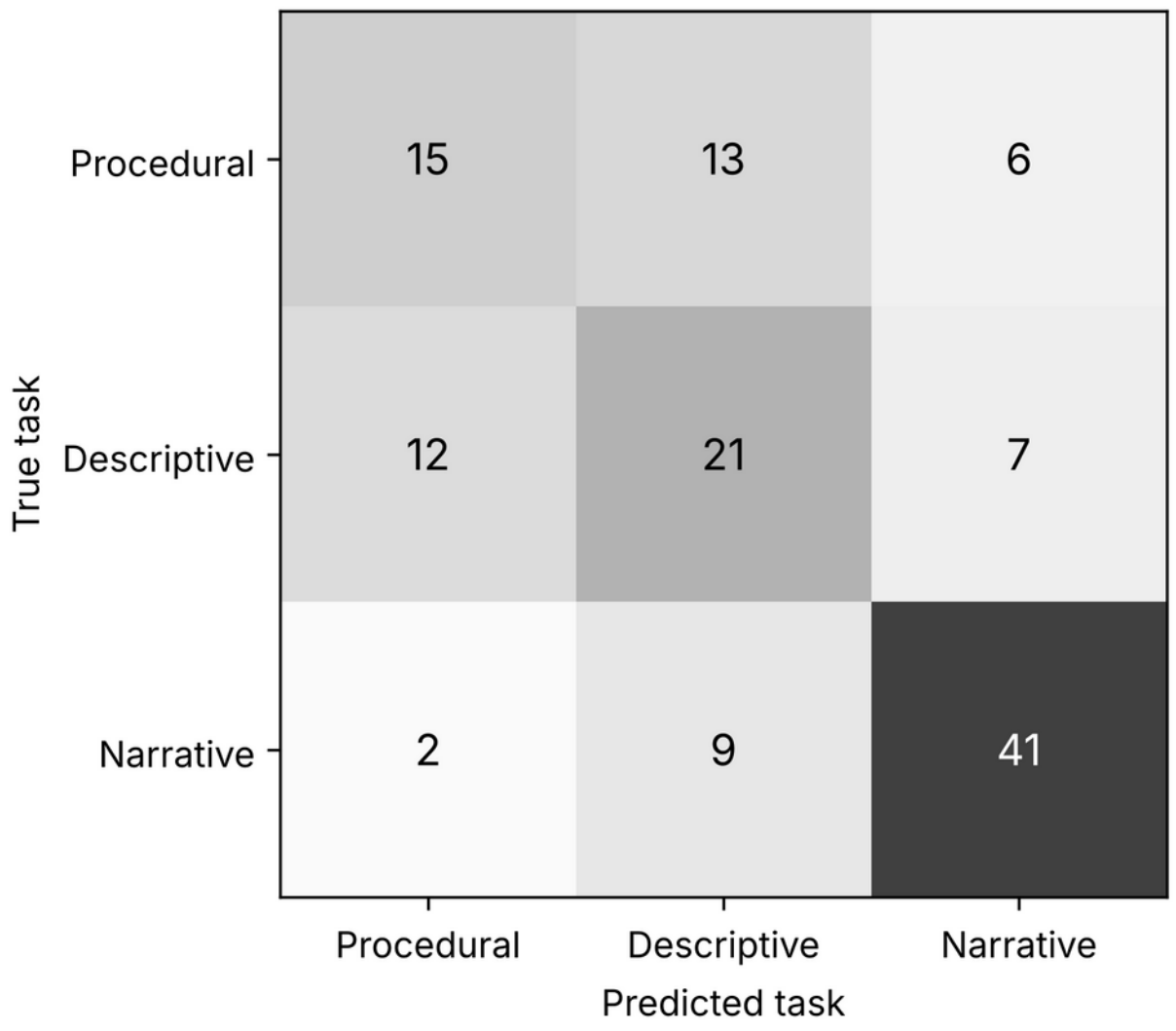

*Note.* Rows show the true task and columns the predicted task. Each of the 126 samples was predicted by a model that had not seen that speaker during training.

The classifier was not the main test of the research question. Its value is that it looks at all features together, and that it was evaluated on speakers it had not seen. Its pattern matches the statistical results: the Narrative task leaves a clear linguistic signature, also when the amount of speech is removed, while the Descriptive and Procedural tasks are hard to tell apart.

## 5. Discussion

This study asked whether the choice of discourse task changes the linguistic measures extracted from aphasia discourse. Within the same speakers, it did. The amount of speech followed the full complexity gradient, but most measures separated the Narrative task from the other two, and the Descriptive and Procedural tasks did not differ on any measure that is independent of sample length.

### 5.1 Finding 1: Discourse Tasks Are Not Interchangeable

The main finding is that discourse tasks are not interchangeable. The same speakers, recorded in the same session, produced systematically different values on most measures depending on the task. A speaker's total word count, NDW or filler rate is therefore not a fixed property of that speaker. It is a joint product of the speaker and the task.

This result agrees with earlier work showing main effects of discourse type on CLAN-derived measures in aphasia (Stark, 2019), genre-specific microstructure in speakers with and without aphasia (Stark & Fukuyama, 2021), only moderate or weak agreement between picture description and conversation in the same speakers (Leaman & Edmonds, 2023), and task effects on almost 300 automatically extracted features across clinical groups (Stark & Themistocleous, 2026). The present study extends this work in three ways, in a design in which every speaker contributed all three tasks. It uses automatically extracted disfluency markers, which are increasingly used in automated assessment. It tests an ordered gradient of task demand, not only an overall difference. And it shows that the task effect on disfluency remains when disfluencies are expressed per word. The classifier pointed in the same direction and recognized Narrative samples most reliably.

### 5.2 Why Narrative Discourse Differs Fundamentally from Procedural Discourse

The largest and most consistent differences were between the Cinderella retell and the Sandwich procedure, and the two tasks differ in at least four ways. First, the Cinderella story must be recalled from memory after the book is removed, and the speaker must hold its macrostructure while producing it episode by episode (Kintsch & van Dijk, 1978); the Sandwich task asks for a short routine that most adults have performed many times. Second, a procedure has a fixed order of steps, each of which can be expressed with a short noun phrase such as "the bread". This gives simple, noun-heavy phrasing and low lexical diversity (Fergadiotis et al., 2011; Stark & Fukuyama, 2021). A story requires the speaker to track characters, mark the order of events and express causes, and it calls for more past tense forms (Riccardi et al., 2024). Third, the story invites long retellings, and in typical speakers disfluency rises with utterance length and with less structured formats (Oviatt, 1995). Fourth, narrative performance in aphasia is linked to executive functioning (Dutta et al., 2024). The Cinderella retell is therefore not simply "more language": it loads memory, planning and monitoring at the same time, which is why it produced the richest and the most disfluent discourse.

### 5.3 Finding 2: Narrative Stands Apart, Even Per Word

The second finding is that the task effect is not a smooth gradient. Only the amount of speech (total word count and NDW) rose step by step, and even there the step from Sandwich to Broken Window was small. On the per-word measures, the Descriptive and Procedural tasks did not differ, while the Narrative task was higher. Three points help to explain this pattern.

**The amount of speech follows the content of the task.** Total words and NDW reflect how much there is to say. The Broken Window pictures show a short sequence of events, which gives slightly more to describe than a sandwich procedure and much less than a full fairy tale. MLU rose only from the Procedural to the Narrative task, so utterance length did not rise steadily with assumed task demand.

**Per-word disfluency rises when the story must be recalled from memory.** Fillers, repetitions and total disfluency density were higher in the Narrative task than in both other tasks, also when expressed per word. Fillers signal planning delays (Clark & Fox Tree, 2002). One explanation is that the Cinderella task is the only task in which the speaker must retrieve the content from memory without pictures. In the Broken Window task the pictures act as external memory, and in this corpus the sandwich task was given with a photo of the sandwich (Section 3.2). The two tasks that did not differ are therefore also the two tasks with visual support, which fits a memory-support account better than a single gradient of demand. A single photo of a sandwich gives less support than a picture sequence, however, so this explanation should be tested directly. This interpretation agrees with Culbertson (2022), who found more repetitions and retracing in the Cinderella story than in picture-sequence descriptions. A similar effect has been reported in healthy adults, who were more disfluent per word when they had to organize their own answer than when they told a story from a picture book (Walters et al., 2026). The revision rate was higher in Narrative than in Procedural discourse but not reliably higher than in Descriptive discourse, and unintelligible speech was higher in Narrative than in Descriptive discourse only, so self-repair and breakdown may be less tied to the memory demands of the task than hesitation is. A simpler explanation should also be considered: disfluency rates were low in the two shorter tasks, and low counts leave little room to detect differences.

**Lexical diversity did not drop in the longer retells.** NDW showed some of the largest task differences in the study, while MATTR, which measures the same property independently of length, showed only a small effect, in the direction of slightly higher diversity in Narrative discourse. The large Narrative advantage in NDW is therefore mainly a length effect, which is the problem that MATTR was designed to solve (Covington & McFall, 2010; Fergadiotis et al., 2015). At the same time, MATTR was not unaffected by task: the retells were slightly more diverse, possibly because a story with several characters and events calls for more varied vocabulary than a short procedure or picture sequence.

### 5.4 Finding 3: Methodological Implications for Future Discourse Research

The results have direct implications for computational and clinical research that extracts linguistic measures from aphasia discourse.

**Discourse task reporting.** Studies that do not report the task make their results difficult to compare. Reporting guidelines for spoken discourse research already ask authors to describe elicitation procedures in detail (Stark et al., 2023). The present results show why this is essential: the same measure can take very different values in different tasks. The task should therefore be reported for every feature and every analysis.

**Pooling samples across tasks.** Combining samples from different tasks into one feature vector, or comparing a feature extracted from one task in one study with the same feature from another task in another study, confounds task with speaker. Bryant et al. (2016) warned that collapsing genres into a single sample may hide difficulties in one genre behind easier performance in another. When several tasks are used, task should be included in the model, for example as a factor or a stratification variable.

**Task-specific reference values.** Because the values depend on the task, reference ranges and cut-offs should be established separately for each task. Efforts to standardize spoken discourse assessment in aphasia are already working toward shared protocols and outcome measures (Stark et al., 2021).

**Length correction.** Raw counts such as total word count and NDW largely track how much a speaker says, and MATTR removed most of the task effect on vocabulary. Per-word disfluency rates, however, still differed between the Narrative task and the other two. Length-corrected measures are therefore better suited than raw counts for comparisons across tasks, but they should still be compared within the same task wherever possible. Raw counts remain useful, because the amount of speech a person produces is itself clinically meaningful.

**Task effects in machine learning.** Task identity can be learned from the features, especially for narrative samples. If different groups of speakers completed different tasks, a model could therefore learn the task instead of the group. Models trained on one task should be tested on other tasks before being described as general, and all samples from a speaker should be kept in the same fold (Saeb et al., 2017).

**Task consistency over time.** In longitudinal studies and treatment trials, a change of task between assessments may look like a change in language ability. The same task should be used at every time point. Repeating the same stimulus has its own risk, because speakers may become familiar with a story over repeated assessments (Gupta et al., 2025). Where this is a concern, parallel stimuli from the same genre may be a better choice than switching genre.

### 5.5 Finding 4: Clinical Implications

The clinical message of this study is simple. Researchers and clinicians who use aphasia discourse should carefully control the elicitation task, because task selection itself influences the distributions of these measures.

In practice, the task should be chosen to match the question and the person's own communication goals (Leaman & Archer, 2023). A Narrative retell such as Cinderella produces the most language and is the most sensitive to breakdowns in planning and fluency, but it is also the most demanding. A Procedural task such as Sandwich is short and familiar and may be easier for people with severe aphasia, but it produces limited and simple language. In this sample, the Broken Window picture sequence gave per-word values similar to the Sandwich procedure, with slightly more speech. These observations agree with the recommendation of Stark and Fukuyama (2021) that a narrative task may be preferable when the goal is to capture the full range of a person's morphosyntactic ability. Narrative discourse also gave the most reliable automated fluency measures over repeated testing in very mild aphasia, while the procedural task gave the least reliable ones (Stark et al., 2026). They also fit current clinical guidance: structured tasks make it possible to elicit similar samples from the same person over time, and procedural discourse may be well suited to measuring speech rate, because it tends to be faster-paced than narration (Dutta et al., 2025).

When discourse is used to monitor a person over time, the same task should be used at each assessment, and a person's values should be compared with reference data from the same task. A higher filler rate in a Cinderella retell than in an earlier Sandwich procedure does not mean that the person's fluency has declined.

### 5.6 Limitations

The main limitations are the following.

- **Fixed task order.** All speakers completed the three tasks in one session and in the same order (Broken Window, Cinderella, Sandwich), so order effects such as fatigue or practice cannot be fully ruled out. A steady order effect would, however, make the measures change gradually from the first task to the last. Instead, the Narrative task, which came in the middle, stood out, while the first and the last task did not differ on any measure that is independent of sample length.
- **Short samples.** The per-word analyses are based on the 31 speakers who produced at least 25 words in every task. Speakers with very short samples, often those with more severe aphasia, are therefore under-represented, and the result for any single disfluency measure should be read with caution.
- **Assumed task demand.** The order Procedural < Descriptive < Narrative is based on theory and was not measured. The tasks also differ in several ways at once, including familiarity, visual support and the amount of content.

- **Dependence on transcription.** The measures rely on disfluency codes inserted by human transcribers, and pauses were coded too rarely to be analyzed.
- **Generalization.** The data come from one corpus of English-speaking adults with chronic post-stroke aphasia, and both the Cinderella story and the sandwich task are culturally specific (Dutta et al., 2025).

### 5.7 Future Directions

Future research should replicate these findings in larger and more varied samples, including other AphasiaBank corpora and speakers without aphasia. Earlier work found task effects in speakers with and without aphasia (Stark, 2019; Stark & Themistocleous, 2026), but whether the pattern found here is specific to aphasia is still open. Other tasks from the AphasiaBank protocol, such as the Cat Rescue and Refused Umbrella pictures and personal narratives, would show whether the Narrative-versus-other pattern holds across a wider range of genres. A direct comparison of the same procedure with and without pictures would test whether visual support explains why the Descriptive and Procedural tasks behaved alike. Mixed-effects count models with speaker as a random effect could include the speakers with short samples and test how task effects interact with aphasia type and severity, since more severe aphasia goes with more similar language across tasks (Stark & Fukuyama, 2021). Finally, as automatic speech recognition increasingly replaces manual transcription, it will be important to test whether the same task effects appear when features are extracted from automatically generated transcripts. This is not yet straightforward. In a recent benchmark of four automatic speech recognition systems on the speech of people with aphasia, the best model had a median word error rate of about 21%, and all four systems under-represented verbs (Kollapally & Akers, 2025). Models that detect disfluencies jointly with speech recognition have been developed for typical spontaneous speech (Futami et al., 2023), but they have not yet been validated for aphasia.

## 6. Conclusion

This study shows that discourse task selection shapes the linguistic measures extracted from aphasia discourse. Within the same speakers, the amount of speech rose from the Procedural to the Descriptive to the Narrative task, but most measures separated the story retell from the two other tasks. Filler and repetition rates and total disfluency density were higher in the Narrative task than in both other tasks, even per word, while length-corrected lexical diversity did not fall and was, if anything, slightly higher. A machine learning classifier recognized Narrative samples well but often confused Descriptive and Procedural samples.

These findings suggest that discourse task is an important methodological variable that should be explicitly reported and controlled in future computational and clinical studies of aphasia, and that correcting for sample length does not make measures from different tasks interchangeable.

## Declarations

### Acknowledgments

The author thanks Julius Fridriksson and colleagues for contributing the Fridriksson-2 corpus to AphasiaBank, and the people with aphasia who took part in the POLAR trial.

### Funding

No funding was obtained for this work.

### Disclosure of Interest

The author reports there are no competing interests to declare.

### Declaration of Generative AI Use

AI assistants (ChatGPT, Claude, and Gemini) were used as supporting tools for literature searches and reference verification, code auditing for potential run-time, implementation and statistical issues, and for language improvement. The research questions, code, analyses, findings, interpretation, and manuscript were developed by the author. All AI-generated suggestions were critically reviewed, and the author takes full responsibility for the accuracy, content, and conclusions of this article.

### Data Availability Statement

The transcripts analyzed in this study belong to the Fridriksson-2 corpus (Fridriksson, 2023; https://doi.org/10.21415/E8FQ-ZY70). They are available to registered members of AphasiaBank (https://aphasia.talkbank.org) under the TalkBank rules for data access and use, which do not allow the author to share them.

### Code Availability

The source code used in this study is available at https://github.com/emmaalkubaisi-star/task-dependent-discourse-measures. It includes the transcript preprocessing, the regex-based feature extraction, the statistical analysis and the machine learning check.